\documentclass[11pt]{article}

\usepackage[american]{babel}
\usepackage{changepage}

\usepackage{natbib} 
\usepackage{mathtools} 
\usepackage{booktabs} 
\usepackage{tikz} 
\usetikzlibrary{shapes, arrows, shapes.geometric, positioning}

\title{Interpretable Latent Variables in Deep State Space Models}

\author{Haoxuan Wu and David S.\ Matteson and Martin T. \ Wells \\
Department of Statistics and Data Science, Cornell University}

\begin{document}
\maketitle

\begin{abstract}
  We introduce a new version of deep state-space models (DSSMs) that combines a recurrent neural network with a state-space framework to forecast time series data. The model estimates the observed series as functions of latent variables that evolve non-linearly through time. Due to the complexity and non-linearity inherent in DSSMs, previous works on DSSMs typically produced latent variables that are very difficult to interpret. Our paper focus on producing interpretable latent parameters with two key modifications. First, we simplify the predictive decoder by restricting the response variables to be a linear transformation of the latent variables plus some noise. Second, we utilize shrinkage priors on the latent variables to reduce redundancy and improve robustness. These changes make the latent variables much easier to understand and allow us to interpret the resulting latent variables as random effects in a linear mixed model. We show through two public benchmark datasets the resulting model improves forecasting performances.
\end{abstract}

\section{Introduction}
Time series forecasting has been an important area of research for long periods of time. With the advancements in technology and ever-looming presence of big data, the importance of producing accurate and interpretable forecasts for large, noisy datasets is higher than ever before. 

A wide variety of methods have previously been utilized for time series forecasting. State-space models (SSMs) have been commonly used for this task \citep{hyndman2008automatic, aoki2013state, hyndman2002state}. SSMs produce interpretable forecasting results and have shown to be every effective at forecasting low-dimensional time-series data \citep{durbin2012time}. However, they tend to struggle in high dimensional settings, be inefficient with large datasets and are prone to high error propagation for long-term forecasting \citep{daum2005nonlinear}. As a result, machine learning based methods such recurrent neural networks (RNN) have been frequently used as alternatives \citep{oancea2014time, connor1994recurrent}. RNNs  have shown to effective at forecasting high dimensional time series. However, a RNN by itself tends to be not enough to learn more intricate patterns in large datasets. 

Deep state-space models (DSSMs) incorporate recurrent neural networks into a state-space framework to provide non-linear mappings of latent variables \citep{gedon2021deep}. In comparison to SSMs, DSSMs provide additionally flexibility in the estimation of the latent variables while maintaining the same general structure. By utilizing recurrent neural networks to learn global patterns across many series, DSSMs have proven to be effective at forecasting large datasets \citep{rangapuram2018advances}. 

At high level, existing machine learning methods are model-agnostic methods or innately interpretable models. The goal of model-agnostic methods is to explain predictions by a black-box approach.  Alternatively, directly interpretable models such as generalized linear models (GLMs) \citep{mccullagh2019generalized} are widely and successfully applied. Despite their effectiveness, one weakness of DSSMs is the lack of interpretability for the latent variables. For many of the existing work \citep{fraccaro2016sequential, chung2015recurrent, krishnan2017structured}, the observed data is assumed to follow a certain probability distribution whose sufficient statistics are calculated by a non-linear, black-box transformation of the latent variables. As a result, it becomes very difficult to explain the relationship between the observed and the latent variables. In this paper, we propose a new version of DSSM that produces interpretable latent variables while improving the forecasting performances. The details of our model will be shown in Section \ref{sec_meth}. We highlight two key contributions as follows.

First, our model restricts the mean of the observed variables to be a linear function of the latent variables. Previous concept, previous works on deep state-space models, especially all those dealing with time series data, have focused on using a neural network for the decoder. We aim to show that we can achieve similar level of results with more interpretability by switching to a linear decoder. With a normal likelihood for the observed variables, our set-up allows the latent variables to be interpreted as random effects in a (generalized) linear mixed model with an exotic variance component \citep{mcculloch2004generalized}. 

Second, we incorporate shrinkage priors into the latent space to reduce redundancy and further enhance interpretability. Shrinkage priors have previous shown to be effective in SSMs \citep{cadonna2020triple, kowal2019dynamic} and in other deep machine learning models  \citep{bhadra2020horseshoe, louizos2017bayesian, ghosh2017model}. However, shrinkage priors, in general, have not been used in deep state-space models, in particular for dealing with latent variables. Our work is a first adaption of these ideas to the much broader deep state-space framework in modeling the latent states. We utilize inverse gamma-gamma shrinkage prior \citep{bai2017inverse} for the latent variables in order to induce more shrinkage toward zero. The resulting increased sparsity makes latent variables easier to understand and interpret. As we will show through two real-world benchmark datasets, incorporating shrinkage priors leads to more robustness and better forecasting performances.

\section{Related Work} \label{sec_rel_wk}
Variational autoencoder, introduced by \citet{kingma2013auto}, produced an efficient method to estimate latent variables that are representative of the observed variables using variational inference. Since then, many works has been done expanding the structure and the application of the framework \citep{bayer2014learning, xue2020deep,  krishnan2017structured, karl2016deep}. \citet{chung2015recurrent} introduced variational recurrent neural network which utilizes high level latent variables to learn structure in sequential data. \citet{fraccaro2016sequential} extended the framework by incorporating a backward recurrent neural network to learn information from future observed variables. \citet{krishnan2015deep} developed Deep Kalman Filter which utilized a non-linear autoregressive structure in the estimation of the latent variables. However, in most of above models, the observed variables are modeled as a non-linear transformation of the latent variables parameterized by neural networks. This set-up makes it very difficult to interpret the meaning of the latent variables.

To induce learning of global structure in the latent variables, three key areas of improvements have been made in recent works. First, simplifying the decoder in the generative model have shown to lead to more information stored in the latent \citep{gulrajani2016pixelvae}. For example, \citet{chen2016variational} utilized noisy version of historical data in the generative process. Second, the loss can be modified to induce more structures in the latent variables. \citet{bowman2015generating} utilized KL-annealing to induce lower weight on the KL-divergence term in the loss function to allow for better learning in the training process. \citet{goyal2017zforcing} added auxiliary variables into the network to ensure the observed can be reconstructed through the latent variables. Third, \citet{rezende2015variational} introduced the idea of normalizing flow which induces a more flexible, scalable class of posterior distributions. \citet{kingma2016improved} and \citet{louizos2017multiplicative} expanded on this idea for better approximate posteriors. Most of these works have been focused on the area of NLP or image recognition. We adopt some of ideas for time series analysis and take it one step further. In addition to ensuring the latent variables contain useful information, our model also ensures these variables are interpretable. 

Within the class of deep state-space models, arguably the most interpretable version is the model introduced by \citet{rangapuram2018advances}. In their model, a RNN estimated the coefficients of a Gaussian state-space model and Kalman Filtering was used to update the latent variables. By utilizing the RNN only for modeling of the coefficients, the model avoided the need for variational inference. While this model maintained a state-space structure, the coefficients themselves may freely change over-time. This makes it difficult to understand the true underlying relationship between the observed and the latent variables. The other work in a DSSM context focused on interpretability is the model introduced by \citep{li2019learning}. The paper utilized automatic relevance determination network to select relevant covariates for each time-step. Unlike their work which focused on interpretability in covariate selection, our work focuses on interpretability in terms of latent variables.

\section{Methodology} \label{sec_meth}
\subsection{Cost Function}
Suppose we have observed time series $\pmb{y}_{1:T}$ with $\{\pmb{y}_t \in R^M\}_{t=1}^T$ and a set of covariates $\pmb{u}_{1:T}$ with $\{\pmb{u}_t \in R^N\}_{t=1}^T$. For our version of the deep state-space model, we assume the observed variables are functions of latent variables $\pmb{z}_{1:T}$ with $\{\pmb{z}_t \in R^Q\}_{t=1}^T$. Suppose further the latent variables depend on a recurrent neural network with output variables $\pmb{h}_{1:T}$, a set of global variables $\pmb{g}$ and a set of local variables $\pmb{\lambda}_{1:T}$. We will detail each of these variables in Subsection \ref{subsec_gen}. For clarity, we will drop subscripts from notation (e.g. $\pmb{y} \equiv \pmb{y}_{1:T}$). Let $\pmb{\theta}$ denote all parameters associated with the DSSM such as weights for the neural networks. We can write the loss as follows:
\begin{align} \label{eq_loss}
L(\pmb{\theta} ; \pmb{y}, \pmb{u}) = L(\pmb{\theta}) 
& \equiv \log \! \int \! p_{\pmb{\theta}} (\pmb{y}, \pmb{z}, \pmb{h},  \pmb{g}, \pmb{\lambda} |  \pmb{u}) d\pmb{z} \, d \pmb{h} \, d\pmb{g} \,  d\pmb{\lambda}
\end{align}
Due to the complexity of the model, this integral is intractable. As a result, we will use an inference model to estimate an approximate posterior $q_{\pmb{\phi}}(\pmb{z}, \pmb{h}, \pmb{g}, \pmb{\lambda} | \pmb{y}, \pmb{u})$ with a set of inference parameters $\pmb{\phi}$. The inference model is utilized to approximate the true posterior $p_{\pmb{\theta}}(\pmb{z}, \pmb{h}, \pmb{g}, \pmb{\lambda} | \pmb{y}, \pmb{u})$. Using Jensen's inequality, Equation \ref{eq_loss} can be bounded by:
\begin{align} \label{eq_vi}
    L(\pmb{\theta}) &\ge E_{q_{\pmb{\phi}}(\cdot)}[\log \frac{p_{\pmb{\theta}}(\pmb{y}, \pmb{z}, \pmb{h},  \pmb{g}, \pmb{\lambda} |  \pmb{u})}{q_{\pmb{\phi}}(\pmb{z}, \pmb{h}, \pmb{g}, \pmb{\lambda} | \pmb{y}, \pmb{u})}] \nonumber \\
    &= E_{q_{\pmb{\phi}}(\cdot)} [\log p_{\pmb{\theta}}(\pmb{y} | \pmb{z}, \pmb{h}, \pmb{g}, \pmb{\lambda}, \pmb{u})]   - \text{KL}(q_{\pmb{\phi}}(\pmb{z}, \pmb{h}, \pmb{g}, \pmb{\lambda} | \pmb{y}, \pmb{u}) \, ||  \, p_{\pmb{\theta}}(\pmb{z}, \pmb{h}, \pmb{g}, \pmb{\lambda} | \pmb{y}, \pmb{u}))
\end{align}
where $q_{\pmb{\phi}}(\cdot) = q_{\pmb{\phi}}(\pmb{z}, \pmb{h}, \pmb{g}, \pmb{\lambda} | \pmb{y}, \pmb{u})$. Equation \ref{eq_vi} is also known as the Variational Evidence Lower Bound (ELBO). The first term is a reconstruction loss composed of the expected log-likelihood. This term encourages the approximate posterior from the inference model to properly explain the observed variables. The second term is the KL-divergence between the prior and the approximate posterior. This term encourages the approximate posterior to be close to the prior. The resulting model can be decomposed into two parts: a generative model to estimate the joint distribution and an inference model to estimate the posterior for the latent variables. 

To ensure the ELBO can be estimated effectively, we design the model (seen in Figure \ref{fig_1}) to allow for closed-form estimation of the KL-divergence term. A closed form solution avoids sampling from $p_{\pmb{\theta}}(\pmb{z}, \pmb{h}, \pmb{g}, \pmb{\lambda} | \pmb{y}, \pmb{u})$ during training and speeds up estimation of the loss. Our choices for the generative/inference portions of the model will be detailed in Subsections \ref{subsec_gen} and \ref{subsec_inf}. A more detailed explanation for the variational inference steps are shown in the Appendix. 

\subsection{Generative Model} \label{subsec_gen}
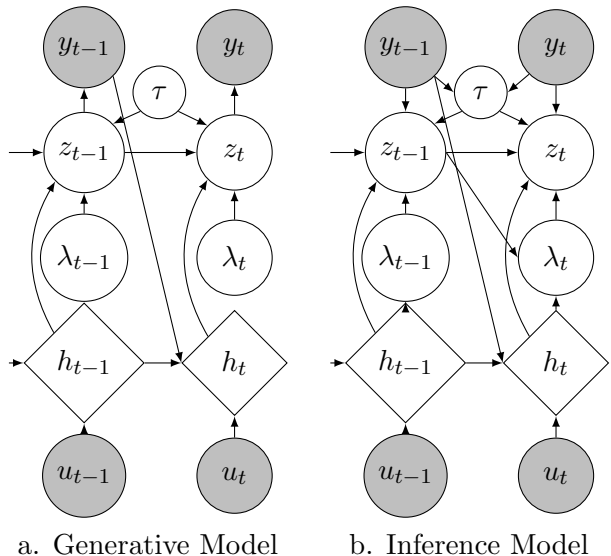
\begin{figure}
\centering
  \setlength{\tabcolsep}{8pt}
  \begin{tabular}{c c}
  \begin{tikzpicture} [ on grid, auto]
\node (h0) [draw, minimum width = 1.2cm,
    minimum height = 1.2cm, diamond] {$h_{t-1}$};
\node (l0) [draw, minimum size=0.9cm, circle, above = 1.4cm of h0] {$\lambda_{t-1}$};
\node (u0) [draw, minimum size=0.9cm, circle, below = 1.5cm of h0, fill=lightgray] {$u_{t-1}$};
\node (z0) [draw, minimum size=0.9cm, circle, above = 1.4cm of l0] {$z_{t-1}$};
\node (y0) [draw, minimum size=0.9cm, circle, above = 1.4cm of z0, fill=lightgray] {$y_{t-1}$};
\node (h1) [draw, minimum width = 1.4cm,
    minimum height = 1.4cm, diamond, right = 2cm of h0] {$h_{t}$};
\node (l1) [draw, minimum size=1.0cm, circle, right = 2cm of l0] {$\lambda_{t}$};
\node (u1) [draw, minimum size=1.0cm, circle, right = 2cm of u0, fill=lightgray] {$u_{t}$};
\node (z1) [draw, minimum size=1.0cm, circle, right = 2cm of z0] {$z_{t}$};
\node (y1) [draw, minimum size=1.0cm, circle, right = 2cm of y0, fill=lightgray]
{$y_{t}$};
\node (tau) [draw, minimum size = 0.7cm, circle, below right = 0.6cm and 1cm of y0] {$\tau$};
\coordinate[left = 1cm of h0] (hinit);
\coordinate[left = 1cm of z0] (zinit);
\draw[-latex]  (hinit) -- (h0.west);
\draw[-latex]  (zinit) -- (z0.west);
\draw[-latex] (h0.north west) to [bend left=30] (z0.south west);
\draw[-latex] (u0.north) -- (h0.south);
\draw[-latex] (l0.north) -- (z0.south);
\draw[-latex] (z0.north) -- (y0.south);
\draw[-latex] (h1.north west) to [bend left=30] (z1.south west);
\draw[-latex] (u1.north) -- (h1.south);
\draw[-latex] (l1.north) -- (z1.south);
\draw[-latex] (z1.north) -- (y1.south);
\draw[-latex] (z0.east) -- (z1.west);
\draw[-latex] (h0.east) -- (h1.west);
\draw[-latex] (y0.south east) -- (h1.west);
\draw[-latex] (tau.south west) -- (z0.north east);
\draw[-latex] (tau.south east) -- (z1.north west);
\end{tikzpicture} & 
\begin{tikzpicture} [ on grid, auto]
\node (h0) [draw, minimum width = 1.2cm,
    minimum height = 1.2cm, diamond] {$h_{t-1}$};
\node (l0) [draw, minimum size=1.0cm, circle, above = 1.4cm of h0] {$\lambda_{t-1}$};
\node (u0) [draw, minimum size=1.0cm, circle, below = 1.5cm of h0, fill=lightgray] {$u_{t-1}$};
\node (z0) [draw, minimum size=1.0cm, circle, above = 1.4cm of l0] {$z_{t-1}$};
\node (y0) [draw, minimum size=1.0cm, circle, above = 1.4cm of z0, fill=lightgray] {$y_{t-1}$};
\node (h1) [draw, minimum width = 1.4cm,
    minimum height = 1.4cm, diamond, right = 2cm of h0] {$h_{t}$};
\node (l1) [draw, minimum size=1.0cm, circle, right = 2cm of l0] {$\lambda_{t}$};
\node (u1) [draw, minimum size=1.0cm, circle, right = 2cm of u0, fill=lightgray] {$u_{t}$};
\node (z1) [draw, minimum size=1.0cm, circle, right = 2cm of z0] {$z_{t}$};
\node (y1) [draw, minimum size=1.0cm, circle, right = 2cm of y0, fill=lightgray] {$y_{t}$};
\node (tau) [draw, minimum size = 0.7cm, circle, below right = 0.6cm and 1cm of y0] {$\tau$};
\coordinate[left = 1cm of h0] (hinit);
\coordinate[left = 1cm of z0] (zinit);
\draw[-latex]  (hinit) -- (h0.west);
\draw[-latex]  (zinit) -- (z0.west);
\draw[-latex] (u0.north) -- (h0.south);
\draw[-latex] (l0.north) -- (z0.south);
\draw[-latex] (h0.north west) to [bend left=30] (z0.south west);
\draw[-latex] (h0.north) -- (l0.south);
\draw[-latex] (u1.north) -- (h1.south);
\draw[-latex] (l1.north) -- (z1.south);
\draw[-latex] (h1.north west) to [bend left=30] (z1.south west);
\draw[-latex] (h1.north) -- (l1.south);
\draw[-latex] (y0.south east) -- (h1.west);
\draw[-latex] (z0.east) -- (z1.west);
\draw[-latex] (h0.east) -- (h1.west);
\draw[-latex] (y0.south) -- (z0.north);
\draw[-latex] (y1.south) -- (z1.north);
\draw[-latex] (z0.east) -- (l1.west);
\draw[-latex] (y0.south east) -- (tau.west);
\draw[-latex] (y1.south west) -- (tau.east);
\draw[-latex] (tau.south west) -- (z0.north east);
\draw[-latex] (tau.south east) -- (z1.north west);
\end{tikzpicture} \\
        a. Generative Model & b. Inference Model \\
    \end{tabular}
    \caption{Details of the processes for the generative and inference portions of the DSSM model. Deterministic components are shown by diamonds; latent variables are shown by unshaded circles and the rest are shown by shaded circles.}
    \label{fig_1}
\end{figure}

The generative portion of the model can be seen in Figure \ref{fig_1}, Part a. The joint distribution can be factorized as follows:
\begin{align} \label{eq_gen}
    p_{\pmb{\theta}}(\pmb{y}, \pmb{z},  \pmb{h}, \pmb{g}, \pmb{\lambda} | \pmb{u}) 
    = p_{\pmb{\theta}}(\pmb{g}) \prod_{t=1}^T &  p_{\pmb{\theta}}( \pmb{y}_t | \pmb{z}_t) p_{\pmb{\theta}}(\pmb{z}_t | \pmb{h}_t, \pmb{z}_{t-1}, \pmb{\lambda}_t, \pmb{g})  p_{\pmb{\theta}}(\pmb{h}_t | \pmb{h}_{t-1}, \pmb{y}_{t-1}, \pmb{u}_t ) p_{\pmb{\theta}}(\pmb{\lambda}_t)
\end{align}
The generative model composed of estimating two global distributions and four time-specific distributions. The global part consists of estimating $\pmb{g}$ while the time-specific part consists of estimating $\pmb{\lambda}_t$ and three conditional distributions. We will detail the choice and the reasoning for each of the distributions below, starting with $p_{\pmb{\theta}}( \pmb{y}_t | \pmb{z}_t)$. Note the initial variables $(\pmb{z}_0, \pmb{y}_0, \pmb{h}_0)$ are omitted from the equations for clarity; they are all initialized to be zero in implementation. 

\subsubsection{Generate Latent Variables}
To allow for more efficient posterior sampling \citep{ingraham2016bayesian}, we write \\
$p_{\pmb{\theta}}(\pmb{z}_t | \pmb{h}_t, \pmb{z}_{t-1}, \pmb{\lambda}_t, \pmb{g})$ in non-centered parameterization form as follows:
\begin{align} \label{eq_gen_z}
    \pmb{z}_t \sim \pmb{z}_t^* \pmb{\tau}_t^* \pmb{\lambda}_t.
\end{align}
The prior for $\pmb{z}_t$ can be decomposed into two parts: a normal prior given by $\pmb{z}_t^*$ and shrinkage components given by $\pmb{\tau}_t^* \pmb{\lambda}_t$. Note that $\pmb{\tau}_t^*$ is a function of both $\pmb{g}$ and $\pmb{\lambda}_t$, details will be shown in Equation \ref{eq_gen_tau_tilde}. 

Note that $\pmb{z}_t^*$ is a function of the output of the recurrent neural network from the current time-step and the latent variables from the previous time-step.
\begin{align} \label{eq_gen_v}
\pmb{z}_t^* \sim N(\pmb{\mu}_{\pmb{\theta}, \pmb{z}}(\pmb{h}_t, \pmb{z}_{t-1}), \pmb{\sigma}_{\pmb{\theta}, \pmb{z}}(\pmb{h}_t, \pmb{z}_{t-1})) 
\end{align}
where $\pmb{\mu}_{\pmb{\theta}, \pmb{z}}(\pmb{h}_t, \pmb{z}_{t-1}) = NN_{\pmb{\theta}, 1}(\pmb{h}_t, \pmb{z}_{t-1})$, $\pmb{\sigma}_{\pmb{\theta}, \pmb{z}}(\pmb{h}_t, \pmb{z}_{t-1})$ = $\text{SoftPlus}(NN_{\pmb{\theta}, 2}(\pmb{h}_t, \pmb{z}_{t-1}))$ in which $NN$ denote a feed-forward neural network. Note we use different numbered subscripts to denote that the feed-forward neural networks are different.

\subsubsection{Generate Shrinkage Variables}
The shrinkage components $\pmb{\tau}_t^*$ and $\pmb{\lambda}_t$ are the regularized global and local components of the shrinkage prior placed on the latent variables.  The purpose of these terms is to reduce redundancy and add additional robustness. We chose inverse gamma-gamma (IG-G) prior \citep{bai2017inverse} because it has strong posterior contraction properties and has previously shown to be effective in a variational inference setting \citep{ghosh2018structured}. The IG-G prior falls under the class of global-local shrinkage priors with a global component to track overall shrinkage and local components to track shrinkage at each time-step. This set-up allows for strong global shrinkage while still maintaining localized adaptivity. We utilize a regularized version of the prior to achieve an bound on the upper tail \citep{piironen2017sparsity}. We will discuss our choice for the local components $\pmb{\lambda}_t$ first followed by the global component. 

Suppose there are $Q$ latent variables estimated at each time-step, the prior models each of the variables $\pmb{\lambda}_t = [\lambda_{t, 1}, ..., \lambda_{t, Q}]$ independently using a standard half-Cauchy distribution denoted by $C^+(0, 1)$. Since approximating the posterior directly from a half-Cauchy distribution can be difficult, we will use the decomposition to write the half-Cauchy as a combination of a Gamma and an inverse-Gamma distribution \citep{neville2014mean}. 
\begin{align} \label{eq_gen_l}
    \lambda_{t, i}^2 &= \alpha_{t, i}\beta_{t, i} & \text{for \;  $i$} = 1, ..., Q \nonumber \\
    \alpha_{t, i} &\sim G(0.5, 1) & \beta_{t, i} \sim IG(0.5, 1) 
\end{align}
where $G(a, b)$ and $IG(a, b)$ is the Gamma and inverse-Gamma distribution with shape parameter $a$ and scale parameter $b$. 

The global component of the model consists of two variables: $\pmb{g} = (\tau, c)$. $\tau$ estimates global amount of shrinkage across all time-steps and $c$ controls the upper bound for the shrinkage components. Details for these components will be given in Equation \ref{eq_gen_tau} and \ref{eq_gen_c}. A similar decomposition can be used for $p_{\pmb{\theta}}(\tau)$. In a IG-G prior, $\tau \sim C^+(0, \tau_0)$ where $\tau_0$ is a hyperparameter. 
\begin{align} \label{eq_gen_tau}
    \tau^2 &= \alpha_{\tau}\beta_{\tau} &  \nonumber \\
    \alpha_{\tau} &\sim G(0.5, \tau_0^2) & \beta_{\tau} \sim IG(0.5, 1).
\end{align}
One potential issue of the half-Cauchy distribution is the fat tail which may lead to undesirable, large values for the latent variables. To bound the upper limit of the shrinkage prior, we write $\pmb{\tau}_t^* = [\tau_{t, 1}^*, ..., \tau_{t, Q}^*]$ as follows:
\begin{align} \label{eq_gen_tau_tilde}
    \tau_{t, i}^{* 2} &= \frac{c^2 \tau^2}{c^2 + \tau^2 \lambda_{t, i}^2} & \text{for \;  $i$} = 1, ..., Q
\end{align}
where $c$ is a weight decay variable for controlling the upper-bound. With this set-up, in time-steps when $\tau^2 \lambda_{t, i}^2 \gg c^2$,  $\tau_{t, i}^{* 2}\lambda_{t, i}^2 \rightarrow c^2$. In time-steps when $\tau^2 \lambda_{t, i}^2 \ll c^2$,  $\tau_{t, i}^{* 2}\lambda_{t, i}^2 \rightarrow \tau^2 \lambda_{t, i}^2$, which is the standard horseshoe. As a result, the regularized version maintains the strong posterior contraction of the horseshoe prior while avoiding arbitrarily high values for the latent variables. As recommended by \citet{piironen2017sparsity}, we place an inverse gamma prior on $c^2$ with two hyperparameters $c_0$ and $c_1$: 
\begin{align} \label{eq_gen_c}
    c^2 \sim IG(c_0, c_1)
\end{align} 

\subsubsection{Generate Response Variables}
As previous discussed in Section 1, we restrict $\{\pmb{y}_t\}_{t=1}^T$ to be a linear transformation of the latent variables plus some noise component. Assuming $\pmb{y}_t$ follows normal distribution, $p_{\pmb{\theta}}( \pmb{y}_t | \pmb{z}_t)$ can be written as follows:
\begin{align} \label{eq_gen_y}
    \pmb{y}_t \sim N(\pmb{A}_{\pmb{\theta}} \pmb{z}_t,  \sigma_{\pmb{\theta}, \pmb{y}}(\pmb{z}_t)) 
\end{align}
where $\sigma_{\pmb{\theta}, \pmb{y}}(\cdot) = \text{SoftPlus}(NN_{\pmb{\theta}, 3}(\pmb{z}_t))$ and $\pmb{A}_{\pmb{\theta}}$ is a linear transformation. We adopt the SoftPlus transformation, previously seen in \citet{li2019learning}, to ensure that the standard deviation is non-negative. We allow the standard deviation to be a non-linear transformation of the latent variables to give additional flexibility for complex noises. Setting $\pmb{A}_{\pmb{\theta}}$ as non-time-varying makes the relationship between the predicted response and the latent variables easier to understand. As we will discuss in Section \ref{sec_lat_int}, this set-up allows $\{\pmb{z}_t\}$ to be interpreted as random effects in a linear mixed model.

\subsection{Inference Model} \label{subsec_inf}
The inference portion of the model can be seen in Figure \ref{fig_1}, Part b. We approximate the true posterior with the following factorization:
\begin{align} \label{eq_inf}
    q_{\pmb{\phi}}(\pmb{z}, \pmb{h},\pmb{g}, \pmb{\lambda} | \pmb{y}, \pmb{u}) = 
    &q_{\pmb{\phi}}(\pmb{g} | \pmb{y}) \prod_{t=1}^T  q_{\pmb{\phi}}(\pmb{z}_t | \pmb{z}_{t-1}, \pmb{y}_t, \pmb{h}_t, \pmb{g}, \pmb{\lambda}_{t})  \nonumber \\
    &q_{\pmb{\phi}}(\pmb{\lambda}_t | \pmb{z}_{t-1}, \pmb{h}_t) p_{\pmb{\theta}} (\pmb{h}_t |  \pmb{h}_{t-1}, \pmb{y}_{t-1}, \pmb{u}_t)
\end{align}
For the inference portion, we estimate the posterior for the global variables $\pmb{g}$ using all available observed variables. Two posterior distributions are calculated at each time-step: an estimation for posterior of $\pmb{z}_t$ and an estimation for the posterior of $\pmb{\lambda}_t$. Since $\pmb{h}_t$ is deterministic, we use the same network as the generative portion. The key to the inference step is finding an appropriate family for the posterior of each variable to allow for estimation of intractable posterior seen in Equation \ref{eq_loss}. We will explain the choice and reasoning for each of the posteriors below, starting with $q_{\pmb{\phi}}(\pmb{z}_t | \pmb{z}_{t-1}, \pmb{y}_t, \pmb{h}_t, \tau, c, \pmb{\lambda}_{t})$.

\subsubsection{Inference of Latent Variables}
Due to the non-centered parameterization for the generative model seen in Equation \ref{eq_gen_z}, we adopt a similar parameterization for the approximate posterior. Note we use tilde notation to indicate the posterior estimates for the variables from the inference network.
\begin{align} \label{eq_inf_z}
    \pmb{\widetilde{z}}_t &\sim \pmb{\widetilde{z}}_t^* \pmb{\widetilde{\tau}}_t^* \pmb{\widetilde{\lambda}}_t \nonumber \\
    \pmb{\widetilde{z}}_t^* &\sim N(\pmb{\mu}_{\pmb{\phi}, \pmb{\widetilde{z}}}(\pmb{\widetilde{z}}_{t-1}, \pmb{y}_t, \pmb{h}_t), \pmb{\sigma}_{\pmb{\phi}, \pmb{\widetilde{z}}}(\pmb{\widetilde{z}}_{t-1}, \pmb{y}_t, \pmb{h}_t))
\end{align}
where $\pmb{\mu}_{\pmb{\phi}, \pmb{\widetilde{z}}}(\pmb{\widetilde{z}}_{t-1}, \pmb{y}_t, \pmb{h}_t) = NN_{\pmb{\phi}, 1}(\pmb{\widetilde{z}}_{t-1}, \pmb{y}_t, \pmb{h}_t)$, ($\pmb{\widetilde{\tau}}_t^*$, $\pmb{\widetilde{\lambda}}_t$) denote the approximate posteriors for the shrinkage variables and $\pmb{\sigma}_{\pmb{\phi}, \pmb{\widetilde{z}}}(\pmb{\widetilde{z}}_{t-1}, \pmb{y}_t, \pmb{h}_t) = \text{SoftPlus}(NN_{\pmb{\phi}, 2}(\pmb{\widetilde{z}}_{t-1}, \pmb{y}_t, \pmb{h}_t))$. With this set-up, the posterior for the latent variables can be expressed as
$$
\pmb{\widetilde{z}}_t \sim N(\pmb{\mu}_{\pmb{\phi}, \pmb{\widetilde{z}}}(\pmb{\widetilde{z}}_{t-1}, \pmb{y}_t, \pmb{h}_t) \pmb{\widetilde{\tau}}_t^* \pmb{\widetilde{\lambda}}_t, \pmb{\sigma}_{\pmb{\phi}, \pmb{\widetilde{z}}}(\pmb{\widetilde{z}}_{t-1}, \pmb{y}_t, \pmb{h}_t) \pmb{\widetilde{\tau}}_t^* \pmb{\widetilde{\lambda}}_t)$$
This allows $\pmb{\widetilde{z}}_t$ to be viewed as a scale mixture of normals with scales $\pmb{\widetilde{\tau}}_t^* \pmb{\widetilde{\lambda}}_t$. 

\subsubsection{Inference of Shrinkage Variables}
From Equation \ref{eq_gen_l}, we see that each of the $\pmb{\widetilde{\lambda}}_t = [\widetilde{\lambda}_{t, 1}, ..., \widetilde{\lambda}_{t, Q}]$ is modeled independently by a gamma and an inverse gamma distribution. We approximate each using a log-normal distribution. We choose log-normal distribution as it has a closed form KL-divergence with Gamma/Inverse-Gamma prior \citep{louizos2017multiplicative} and allows for efficient sampling \citep{ghosh2017model}. The approximate posterior looks as follows:
\begin{align} \label{eq_inf_lambda}
    \widetilde{\lambda}_{t, i}^2 &= \widetilde{\alpha}_{t, i} \widetilde{\beta}_{t, i} \; \; \; \; \; \; \; \; \; \; \; \; \; \; \; \; \; \; \; \; \; \text{for $i$} = 1, ..., Q  \nonumber \\
    \widetilde{\alpha}_{t, i} &\sim LN(\mu_{\pmb{\phi}, \alpha, t, i}(\pmb{z}_{t-1},  \pmb{h}_t), \sigma_{\pmb{\phi}, \alpha, t, i}(\pmb{z}_{t-1}, \pmb{h}_t)) \nonumber \\ 
    \widetilde{\beta}_{t, i} &\sim LN(\mu_{\pmb{\phi}, \beta, t, i}(\pmb{z}_{t-1}, \pmb{h}_t), \sigma_{\pmb{\phi}, \beta, t, i}(\pmb{z}_{t-1}, \pmb{h}_t))
\end{align}
where $LN(\cdot, \cdot)$ represents a log-normal distribution. The posterior parameters \\
$(\mu_{\pmb{\phi}, \alpha, t, i}(\pmb{z}_{t-1}, \pmb{h}_t), \sigma_{\pmb{\phi}, \alpha, t, i}(\pmb{z}_{t-1}, \pmb{h}_t)$, $\mu_{\pmb{\phi}, \beta, t, i}(\pmb{z}_{t-1},  \pmb{h}_t)$, $\sigma_{\pmb{\phi}, \beta, t, i}(\pmb{z}_{t-1}, \pmb{h}_t))$ are each estimated \\
through a feed-forward neural network. The posterior for the global portion is shown in the Appendix. 

\section{Latent Variables Interpretability} \label{sec_lat_int}
As previously mentioned, directly interpretable models such as generalized linear models (GLMs) \citep{mccullagh2019generalized} have been widely and successfully applied.  Equation \ref{eq_gen_y} shows the relationship between the response and the latent variables. From the set-up, it's clear that $\{\pmb{A}_{\pmb{\theta}}\pmb{z}_t\}$ is the linear predictor component of the normal likelihood. By utilizing a linear layer rather than a deep neural network in the generative framework, the model makes the relationship between the response and the latent variables much easier to understand, therefore enhancing the interpretability of the latent variables. Furthermore, shrinkage priors are utilized to borrow strength across the variables \citep{fourdrinier2018shrinkage, seto2021halo} in order to reduce redundancy in the latent variables. In particular, from Equations \ref{eq_gen_z} - \ref{eq_gen_y} one can see that the latent variables can be interpreted as random effects in a linear mixed model with an exotic variance component. 

\subsection{Illustrate Example}
For an illustrative example of the resulting latent variables from our version of deep state-space model, we simulate data from a linear state-space model as follows:
\[
    y_t = \begin{bmatrix}
    1 \\
    0.5
    \end{bmatrix} \pmb{\beta}_t + \epsilon_t \; \; \; \; \; \; \; \; \; \; \; \;  \; \; \; \; \; \; \;  \; \; \; \; \; \; \;  \; \; \; \; \; \; \epsilon_t \sim_{\text{ind}} N(0, 1)
\]
\[
    \pmb{\beta}_t = \begin{bmatrix}
    0.7 & 0.8 \\
    0.0 & 0.9
    \end{bmatrix} \pmb{\beta}_{t-1} + \begin{bmatrix}
    -1.0 \\
    0.9
    \end{bmatrix} u_t +  \eta_t \; \; \; \; \; \; \eta_t \sim_{\text{ind}} N(0, 0.25)
\]
The covariates $u_t \sim U(-1, 1)$. This simple data generation process has previously been tested as a benchmark for various kinds of deep state-space models \citep{gedon2021deep} and can illustrate a direct comparison between the predicted latent variables versus true latent variables. We generate 2560 training samples and 128 test samples of length 100; we ran our version of DSSM on the above data.

\begin{figure*} 
  \centering
  \setlength{\tabcolsep}{0pt}
  \begin{tabular}{c c}
      $\{y_t\}$ & $\{\beta_{t, 1}\}$  \\
      \includegraphics[width=0.4\linewidth]{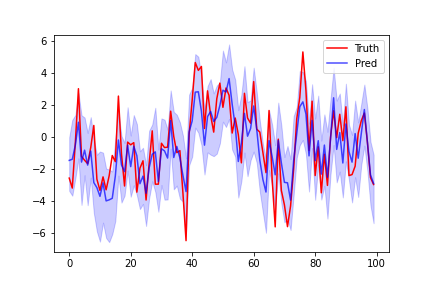}  & 
      \includegraphics[width=0.4\linewidth]{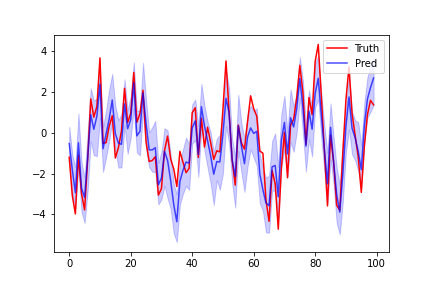}  \\
      $\{\beta_{t, 2}\}$ & Recovery Rate\\
      \includegraphics[width=0.4\linewidth]{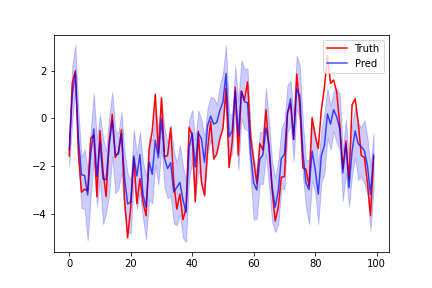}  &
      \includegraphics[width=0.4\linewidth]{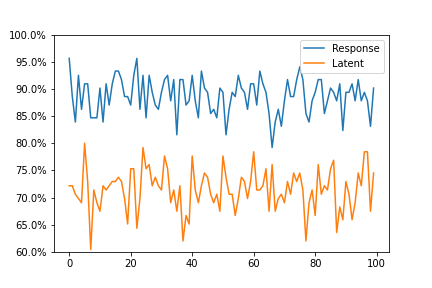} 
  \end{tabular}
   \caption{Top left plot shows example of simulated response $\{y_t\}$ in red, with predicted values from our version of DSSM in blue. The blue bands reflect 90$\%$ credible intervals for sampled forecast values. The top right and bottom left shows the same results for the latent variables $\{\beta_{t,1}\}$ and $\{\beta_{t, 2}\}$. Bottom right plots shows the average recovery rate of the credible intervals for both response and the latent variables across all time-steps.}
   \label{fig_sim_ex}
\end{figure*}

Figure \ref{fig_sim_ex} shows an example of the results from model on the simulated data. As seen in the figure, our version of the deep state-space model is able to produce accurate long-term forecasts of both the response variables and the latent variables. The $90\%$ credible bands reflect inherent noise in the underlying data and can capture a high percentage of true points. We achieve an an average recovery rate of $88.4\%$ for the response and $70.3\%$ for the latent variables that's consistent over the entire forecasting horizon. This highlights the advantage of our model in its ability to recover the true latent variables. By simplifying the relationship between the response and the latent variables, we can induce latent results that can be interpreted as random effects. 

\section{Real Data Analysis}
We utilize two datasets for our analysis: \textit{electricity} and \textit{traffic}. Both datasets are publicly available under the UCI machine learning directory. \textit{Electricity} contains electricity usage of 370 customers, aggregated at an hourly level. \textit{Traffic} contains occupancy rate between 0 and 1 of 440 car lanes on the San Francisco freeways, also aggregated at an hourly level. These 2 datasets represent challenging forecasting problems as they are large datasets containing long-term trends and daily/weekly seasonal fluctuations. These datasets have also been evaluated in previous work \citep{rangapuram2018advances, salinas2020deepar}, making them good benchmarks for comparisons.

Throughout the simulations, we will refer the model introduced in Section 3 as deep state-space model with shrinkage (DSSM-SH). We will compare against 3 competitors: ARIMA, DeepAR \citep{salinas2020deepar} and DSSM \citep{rangapuram2018advances}. ARIMA will act as a state-space model benchmark for these datasets; we utilize the auto.arima function R package \textit{forecast} for automatic selection of orders. DeepAR utilizes a recurrent neural network for prediction. DeepAR has less components on top of the RNN compared to DSSM-SH, making it a good machine learning benchmark for comparison. DSSM, as previously mentioned in Section \ref{sec_rel_wk}, is another deep state-space model. However, unlike DSSM-SH, it utilizes RNN to calculate the coefficients of the state-space framework rather than the latent variables themselves. 

Two metrics will be used for comparisons: normalised deviation (ND) and normalised root mean squared error (RMSE). Details about the metrics, training/testing sets and hyperparameter tuning are documented in the Appendix. 

\subsection{Electricity/Traffic Results}
The results for the models can be seen in Table 1. As seen in the table, DSSM-SH outperformed all competitors in terms both of normalised deviation and normalised root mean squared error on both datasets. This indicates that DSSM-SH does not sacrifice forecasting accuracy in return for interpretable latent variables. Looking at the competing methods, ARIMA struggles with long term forecasting as reflected by the high RMSE. The gap between DSSM-SH and DeepAR highlights the benefits of using a deep state-space framework in comparison to just using a recurrent neural network. Comparing our model with DSSM, we can see that having shrinkage priors in a deep state-space framework leads to better performance. Despite simplifying the expected value of observed variables to be a linear function of the latent variables, DSSM-SH still manages to improve the forecasting results. 

\begin{table*}
\label{tab_nd}
\centering
\caption{Forecasting comparison across 4 models in terms of normalised deviation and normalised root mean squared error for the \textit{electricity} and \textit{traffic} dataset. Error bars across 4 random seeds are shown in subscripts.}
\begin{tabular}{ c | c | c c c c}
\hline 
  & & ARIMA & DeepAR & DSSM & DSSM-SH \\
   \hline
  Electricity & ND & $0.342_{\pm 0.001}$ & $0.079_{\pm 0.003}$ & $0.082_{\pm 0.008}$ & $\pmb{0.071}_{\pm 0.005}$ \\
  & RMSE & $0.892_{\pm 0.002}$ & $0.661_{\pm 0.012}$ & $0.674_{\pm 0.023}$ & $\pmb{0.502}_{\pm 0.035}$ \\ 
  \hline
  Traffic & ND & $0.372_{\pm 0.001}$ & $0.178_{\pm 0.007}$ & $0.185_{\pm 0.018}$ & $\pmb{0.125}_{\pm 0.013}$ \\
  & RMSE & $1.324_{\pm 0.004}$ & $0.433_{\pm 0.009}$ & $0.456_{\pm 0.015}$ & $\pmb{0.373}_{\pm 0.021}$ \\
  \hline 
\end{tabular}
\end{table*}

\begin{figure*} 
  \centering
  \setlength{\tabcolsep}{0pt}
  \begin{tabular}{c c}
  Electricity & Traffic \\
    \includegraphics[width=0.4\linewidth]{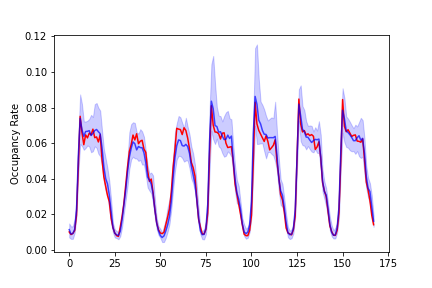} &
    \includegraphics[width=0.4\linewidth]{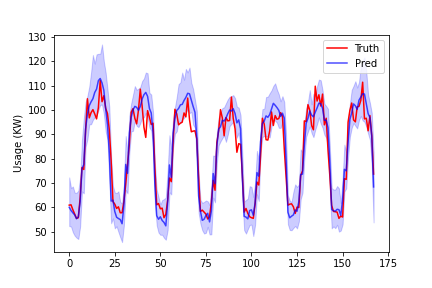}\\
    \end{tabular}
   \caption{Plots shows 1 week rolling forecast from DSSM-SH of a series from the \textit{electricity} dataset and a series from the \textit{traffic} dataset over test horizon of 1 week. x-axis indicate number of hours ahead in the forecasting horizon. }
   \label{fig_traf_for}
\end{figure*}

Figure \ref{fig_traf_for} illustrates forecasting result from DSSM-SH for a series in \textit{electricity} and a series in \textit{traffic}. The plot is generated using a rolling forecast \citep{nicholson2017varx} with a window of 48 hours across the one week testing horizon.  As seen in the results, DSSM-SH is able to capture the long-term patterns inherent in the data and make accurate prediction throughout the forecasting horizon. 50 samples are generated at each forcasting step; the blue bands seen in the figure reflects 95th percentile and 5th percentile of the samples. In time-steps with high occupancy rate, we see that the band becomes wider, reflecting the fact that there exist more volatility in the underlying data. In time-steps where the occupancy rate starts to decrease to a lower level, the band becomes much narrower, reflecting a higher confidence from DSSM-SH about the underlying value. This shows that DSSM-SH is able to capture well the true underlying mean and volatility throughout the forecasting horizon.

\subsection{Ablation Analysis}
To highlight the importance of the shrinkage priors and the linear decoder, we added two ablation analyses. First, to evaluate the importance of shrinkage variables, we performed a comparative experiment with 2 settings. For the first setting, we randomly removed a portion of shrinkage variables by ignoring them during testing time and evaluated the resulting impact on error rate during forecasting. For the second setting, we threshold a percentage of the latent variables with the lowest shrinkage parameter values to 0 and evaluated the resulting error rates. For both experiments, we chose 4 different sparsity percentage: $\{5\%, 10\%, 25\%, 50\%\}$ and calculated the percent error increase. 

Second, to illustrate the effectiveness of the linear layer, we trained two models under same setting, the first using a linear decoder and the second using a neural network for the decoder. We then set a portion of the lowest latent variables whose magnitude is below a certain threshold to 0 and evaluated the impact on error rates. As before, we chose thresholds to induce 4 different sparsity settings of $\{5\%, 10\%, 25\%, 50\%\}$,

\begin{figure}
\setlength{\tabcolsep}{1pt}
\centering
    \begin{tabular}{c c}
        Electricity & Traffic \\
        \includegraphics[width=0.4\linewidth]{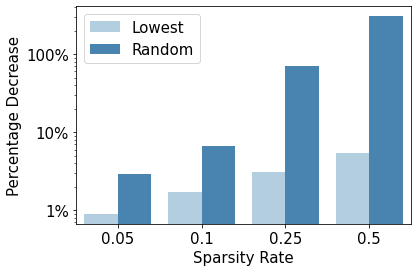} & \includegraphics[width=0.4\linewidth]{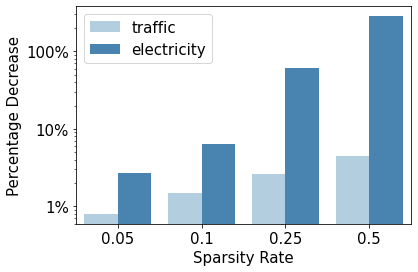} \\
        \includegraphics[width=0.4\linewidth]{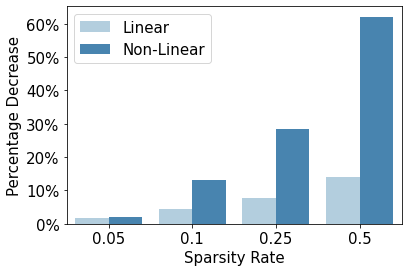} & \includegraphics[width=0.4\linewidth]{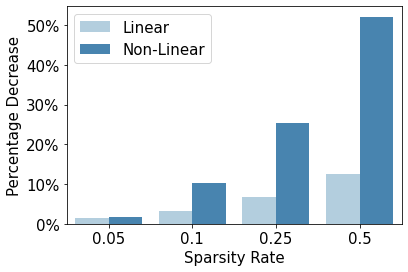} \\
    \end{tabular}
    \caption{The top plots show comparative increases in error rates as a result of randomly removing a portion of shrinkage variables versus setting the latent variables with lowest shrinkage values to 0. The bottom plots show results of setting lowest latent variables to 0 for linear versus non-linear decoders.}
    \label{fig_abla}
\end{figure}

The results are shown in Figure \ref{fig_abla}. For the first analysis pertaining to shrinkage priors, the error increase the two settings are widely different. In the first case of randomly removing shrinkage variables, the error increased significantly, especially when a over $25\%$ of the variables are removed. In the second case, the results did not increase much even when the lowest $50\%$ of latent variables are set to 0. This shows that importance of shrinkage priors in inducing additional robustness through the latent variables. The shrinkage variables can correctly identify unimportant latent variables and shrink them toward 0.

For the second analysis, the linear decoder induces a much clearer relationship between the response and the latent variables. Setting the lowest $50\%$ of latent variables to 0 only decreases forecasting error by $4.9\%$. The magnitudes of the latent variables have a direct correspondence with their importance in the model, supporting the idea that they can be interpreted as random effects. In comparison, the relationship for the non-linear decoder is less clear. 

From the two ablation studies, we highlight how the two primary contributions of the proposed model work together. The linear decoder simplifies the relationship between the response and the latent variables, and the shrinkage priors push insignificant latent variables toward 0. Together they induce an easier interpretation of the latent state.

\section{Conclusion}
In this paper, we introduced a deep state-space model (DSSM-SH) with the ability to produce interpretable latent variables. Our two key modifications to the DSSM framework are restricting the relationship between the response and the latent variables to be a linear function plus some noise component, and incorporating inverse gamma-gamma priors for the latent variables. As we have shown through two real world benchmark datasets, these modifications lead to more interpretable latent results and improved forecasting performances. Our model outperforms both ARIMA (a state-space model) and DeepAR (a RNN based model), showing the effectiveness of the DSSM framework in combining the two parts. In comparison to DSSM, a deep state-space model without shrinkage, our model performs better in forecasting metrics. This shows that add shrinkage priors reduce redundancy and improve robustness.

\bibliography{paper}

\begin{thebibliography}{43}
\providecommand{\natexlab}[1]{#1}
\providecommand{\url}[1]{\texttt{#1}}
\expandafter\ifx\csname urlstyle\endcsname\relax
  \providecommand{\doi}[1]{doi: #1}\else
  \providecommand{\doi}{doi: \begingroup \urlstyle{rm}\Url}\fi

\bibitem[Aoki(2013)]{aoki2013state}
Masanao Aoki.
\newblock \emph{State space modeling of time series}.
\newblock Springer Science \& Business Media, 2013.

\bibitem[Bai and Ghosh(2017)]{bai2017inverse}
Ray Bai and Malay Ghosh.
\newblock The inverse gamma-gamma prior for optimal posterior contraction and
  multiple hypothesis testing.
\newblock \emph{arXiv preprint arXiv:1710.04369}, 2017.

\bibitem[Bayer and Osendorfer(2014)]{bayer2014learning}
Justin Bayer and Christian Osendorfer.
\newblock Learning stochastic recurrent networks.
\newblock \emph{arXiv preprint arXiv:1411.7610}, 2014.

\bibitem[Bhadra et~al.(2020)Bhadra, Datta, Li, and Polson]{bhadra2020horseshoe}
Anindya Bhadra, Jyotishka Datta, Yunfan Li, and Nicholas Polson.
\newblock Horseshoe regularisation for machine learning in complex and deep
  models.
\newblock \emph{International Statistical Review}, 88\penalty0 (2):\penalty0
  302--320, 2020.

\bibitem[Bowman et~al.(2015)Bowman, Vilnis, Vinyals, Dai, Jozefowicz, and
  Bengio]{bowman2015generating}
Samuel~R Bowman, Luke Vilnis, Oriol Vinyals, Andrew~M Dai, Rafal Jozefowicz,
  and Samy Bengio.
\newblock Generating sentences from a continuous space.
\newblock \emph{arXiv preprint arXiv:1511.06349}, 2015.

\bibitem[Cadonna et~al.(2020)Cadonna, Fr{\"u}hwirth-Schnatter, and
  Knaus]{cadonna2020triple}
Annalisa Cadonna, Sylvia Fr{\"u}hwirth-Schnatter, and Peter Knaus.
\newblock Triple the gamma—a unifying shrinkage prior for variance and
  variable selection in sparse state space and tvp models.
\newblock \emph{Econometrics}, 8\penalty0 (2):\penalty0 20, 2020.

\bibitem[Chen et~al.(2016)Chen, Kingma, Salimans, Duan, Dhariwal, Schulman,
  Sutskever, and Abbeel]{chen2016variational}
Xi~Chen, Diederik~P Kingma, Tim Salimans, Yan Duan, Prafulla Dhariwal, John
  Schulman, Ilya Sutskever, and Pieter Abbeel.
\newblock Variational lossy autoencoder.
\newblock \emph{arXiv preprint arXiv:1611.02731}, 2016.

\bibitem[Chung et~al.(2015)Chung, Kastner, Dinh, Goel, Courville, and
  Bengio]{chung2015recurrent}
Junyoung Chung, Kyle Kastner, Laurent Dinh, Kratarth Goel, Aaron~C Courville,
  and Yoshua Bengio.
\newblock A recurrent latent variable model for sequential data.
\newblock \emph{Advances in neural information processing systems},
  28:\penalty0 2980--2988, 2015.

\bibitem[Connor et~al.(1994)Connor, Martin, and Atlas]{connor1994recurrent}
Jerome~T Connor, R~Douglas Martin, and Les~E Atlas.
\newblock Recurrent neural networks and robust time series prediction.
\newblock \emph{IEEE transactions on neural networks}, 5\penalty0 (2):\penalty0
  240--254, 1994.

\bibitem[Daum(2005)]{daum2005nonlinear}
Fred Daum.
\newblock Nonlinear filters: beyond the kalman filter.
\newblock \emph{IEEE Aerospace and Electronic Systems Magazine}, 20\penalty0
  (8):\penalty0 57--69, 2005.

\bibitem[Durbin and Koopman(2012)]{durbin2012time}
James Durbin and Siem~Jan Koopman.
\newblock \emph{Time series analysis by state space methods}.
\newblock Oxford university press, 2012.

\bibitem[Fourdrinier et~al.(2018)Fourdrinier, Strawderman, and
  Wells]{fourdrinier2018shrinkage}
Dominique Fourdrinier, William~E Strawderman, and Martin~T Wells.
\newblock \emph{Shrinkage estimation}.
\newblock Springer, 2018.

\bibitem[Fraccaro et~al.(2016)Fraccaro, S{\o}nderby, Paquet, and
  Winther]{fraccaro2016sequential}
Marco Fraccaro, S{\o}ren~Kaae S{\o}nderby, Ulrich Paquet, and Ole Winther.
\newblock Sequential neural models with stochastic layers.
\newblock \emph{arXiv preprint arXiv:1605.07571}, 2016.

\bibitem[Gedon et~al.(2021)Gedon, Wahlstrom, Schon, and Ljung]{gedon2021deep}
Daniel Gedon, Niklas Wahlstrom, Thomas~B. Schon, and Lennart Ljung.
\newblock Deep state space models for nonlinear system identification.
\newblock \emph{IFAC-PapersOnLine}, 54:\penalty0 481--486, 2021.

\bibitem[Ghosh and Doshi-Velez(2017)]{ghosh2017model}
Soumya Ghosh and Finale Doshi-Velez.
\newblock Model selection in bayesian neural networks via horseshoe priors.
\newblock \emph{arXiv preprint arXiv:1705.10388}, 2017.

\bibitem[Ghosh et~al.(2018)Ghosh, Yao, and Doshi-Velez]{ghosh2018structured}
Soumya Ghosh, Jiayu Yao, and Finale Doshi-Velez.
\newblock Structured variational learning of bayesian neural networks with
  horseshoe priors.
\newblock In \emph{International Conference on Machine Learning}, pages
  1744--1753. PMLR, 2018.

\bibitem[Goyal et~al.(2017)Goyal, Sordoni, C{\^o}t{\'e}, Ke, and
  Bengio]{goyal2017zforcing}
Anirudh Goyal, Alessandro Sordoni, Marc-Alexandre C{\^o}t{\'e}, Nan~Rosemary
  Ke, and Yoshua Bengio.
\newblock Z-forcing: Training stochastic recurrent networks.
\newblock In \emph{NIPS}, 2017.

\bibitem[Gulrajani et~al.(2016)Gulrajani, Kumar, Ahmed, Taiga, Visin, Vazquez,
  and Courville]{gulrajani2016pixelvae}
Ishaan Gulrajani, Kundan Kumar, Faruk Ahmed, Adrien~Ali Taiga, Francesco Visin,
  David Vazquez, and Aaron Courville.
\newblock Pixelvae: A latent variable model for natural images.
\newblock \emph{arXiv preprint arXiv:1611.05013}, 2016.

\bibitem[Hyndman and Khandakar(2008)]{hyndman2008automatic}
Rob~J Hyndman and Yeasmin Khandakar.
\newblock Automatic time series forecasting: the forecast package for r.
\newblock \emph{Journal of statistical software}, 27\penalty0 (1):\penalty0
  1--22, 2008.

\bibitem[Hyndman et~al.(2002)Hyndman, Koehler, Snyder, and
  Grose]{hyndman2002state}
Rob~J Hyndman, Anne~B Koehler, Ralph~D Snyder, and Simone Grose.
\newblock A state space framework for automatic forecasting using exponential
  smoothing methods.
\newblock \emph{International Journal of forecasting}, 18\penalty0
  (3):\penalty0 439--454, 2002.

\bibitem[Ingraham and Marks(2016)]{ingraham2016bayesian}
John~B Ingraham and Debora~S Marks.
\newblock Bayesian sparsity for intractable distributions.
\newblock \emph{arXiv preprint arXiv:1602.03807}, 2016.

\bibitem[Karl et~al.(2016)Karl, Soelch, Bayer, and Van~der Smagt]{karl2016deep}
Maximilian Karl, Maximilian Soelch, Justin Bayer, and Patrick Van~der Smagt.
\newblock Deep variational bayes filters: Unsupervised learning of state space
  models from raw data.
\newblock \emph{arXiv preprint arXiv:1605.06432}, 2016.

\bibitem[Kingma and Welling(2013)]{kingma2013auto}
Diederik~P Kingma and Max Welling.
\newblock Auto-encoding variational bayes.
\newblock \emph{arXiv preprint arXiv:1312.6114}, 2013.

\bibitem[Kingma et~al.(2016)Kingma, Salimans, Jozefowicz, Chen, Sutskever, and
  Welling]{kingma2016improved}
Durk~P Kingma, Tim Salimans, Rafal Jozefowicz, Xi~Chen, Ilya Sutskever, and Max
  Welling.
\newblock Improved variational inference with inverse autoregressive flow.
\newblock \emph{Advances in neural information processing systems},
  29:\penalty0 4743--4751, 2016.

\bibitem[Kowal et~al.(2019)Kowal, Matteson, and Ruppert]{kowal2019dynamic}
Daniel~R Kowal, David~S Matteson, and David Ruppert.
\newblock Dynamic shrinkage processes.
\newblock \emph{Journal of the Royal Statistical Society: Series B (Statistical
  Methodology)}, 81\penalty0 (4):\penalty0 781--804, 2019.

\bibitem[Krishnan et~al.(2017)Krishnan, Shalit, and
  Sontag]{krishnan2017structured}
Rahul Krishnan, Uri Shalit, and David Sontag.
\newblock Structured inference networks for nonlinear state space models.
\newblock In \emph{Proceedings of the AAAI Conference on Artificial
  Intelligence}, volume~31, 2017.

\bibitem[Krishnan et~al.(2015)Krishnan, Shalit, and Sontag]{krishnan2015deep}
Rahul~G Krishnan, Uri Shalit, and David Sontag.
\newblock Deep kalman filters.
\newblock \emph{arXiv preprint arXiv:1511.05121}, 2015.

\bibitem[Li et~al.(2019)Li, Yan, Yang, and Jin]{li2019learning}
Longyuan Li, Junchi Yan, Xiaokang Yang, and Yaohui Jin.
\newblock Learning interpretable deep state space model for probabilistic time
  series forecasting.
\newblock \emph{Proceedings of the Twenty-Eighth International Joint Conference
  on Artificial Intelligence}, 2019.

\bibitem[Lim et~al.(2021)Lim, Arik, Loeff, and Pfister]{lim2021temporal}
Bryan Lim, Sercan Arik, Nicholas Loeff, and Tomas Pfister.
\newblock Temporal fusion transformers for interpretable multi-horizon time
  series forecasting.
\newblock In \emph{International Journal of Forecasting}, 2021.

\bibitem[Louizos and Welling(2017)]{louizos2017multiplicative}
Christos Louizos and Max Welling.
\newblock Multiplicative normalizing flows for variational bayesian neural
  networks.
\newblock In \emph{International Conference on Machine Learning}, pages
  2218--2227. PMLR, 2017.

\bibitem[Louizos et~al.(2017)Louizos, Ullrich, and
  Welling]{louizos2017bayesian}
Christos Louizos, Karen Ullrich, and Max Welling.
\newblock Bayesian compression for deep learning.
\newblock \emph{Advances in neural information processing systems}, 30, 2017.

\bibitem[McCullagh and Nelder(2019)]{mccullagh2019generalized}
Peter McCullagh and John~A Nelder.
\newblock \emph{Generalized linear models}.
\newblock Routledge, 2019.

\bibitem[McCulloch and Searle(2004)]{mcculloch2004generalized}
Charles~E McCulloch and Shayle~R Searle.
\newblock \emph{Generalized, linear, and mixed models}.
\newblock John Wiley \& Sons, 2004.

\bibitem[Neville et~al.(2014)Neville, Ormerod, and Wand]{neville2014mean}
Sarah~E Neville, John~T Ormerod, and MP~Wand.
\newblock Mean field variational bayes for continuous sparse signal shrinkage:
  pitfalls and remedies.
\newblock \emph{Electronic Journal of Statistics}, 8\penalty0 (1):\penalty0
  1113--1151, 2014.

\bibitem[Nicholson et~al.(2017)Nicholson, Matteson, and
  Bien]{nicholson2017varx}
William~B Nicholson, David~S Matteson, and Jacob Bien.
\newblock Varx-l: Structured regularization for large vector autoregressions
  with exogenous variables.
\newblock \emph{International Journal of Forecasting}, 33\penalty0
  (3):\penalty0 627--651, 2017.

\bibitem[Oancea and Ciucu(2014)]{oancea2014time}
Bogdan Oancea and {\c{S}}tefan~Cristian Ciucu.
\newblock Time series forecasting using neural networks.
\newblock \emph{arXiv preprint arXiv:1401.1333}, 2014.

\bibitem[Piironen and Vehtari(2017)]{piironen2017sparsity}
Juho Piironen and Aki Vehtari.
\newblock Sparsity information and regularization in the horseshoe and other
  shrinkage priors.
\newblock \emph{Electronic Journal of Statistics}, 11\penalty0 (2):\penalty0
  5018--5051, 2017.

\bibitem[Rangapuram et~al.(2018)Rangapuram, Seeger, Gasthaus, Stella, Wang, and
  Januschowski]{rangapuram2018advances}
Syama~Sundar Rangapuram, Matthias~W Seeger, Jan Gasthaus, Lorenzo Stella,
  Yuyang Wang, and Tim Januschowski.
\newblock Deep state space models for time series forecasting.
\newblock In \emph{Advances in Neural Information Processing Systems},
  volume~31. Curran Associates, Inc., 2018.

\bibitem[Rezende and Mohamed(2015)]{rezende2015variational}
Danilo Rezende and Shakir Mohamed.
\newblock Variational inference with normalizing flows.
\newblock In \emph{International conference on machine learning}, pages
  1530--1538. PMLR, 2015.

\bibitem[Rezende et~al.(2014)Rezende, Mohamed, and
  Wierstra]{rezende2014stochastic}
Danilo~Jimenez Rezende, Shakir Mohamed, and Daan Wierstra.
\newblock Stochastic backpropagation and approximate inference in deep
  generative models.
\newblock In \emph{International conference on machine learning}, pages
  1278--1286. PMLR, 2014.

\bibitem[Salinas et~al.(2020)Salinas, Flunkert, Gasthaus, and
  Januschowski]{salinas2020deepar}
David Salinas, Valentin Flunkert, Jan Gasthaus, and Tim Januschowski.
\newblock Deepar: Probabilistic forecasting with autoregressive recurrent
  networks.
\newblock \emph{International Journal of Forecasting}, 36:\penalty0 1181--1191,
  2020.

\bibitem[Seto et~al.(2021)Seto, Wells, and Zhang]{seto2021halo}
Skyler Seto, Martin~T Wells, and Wenyu Zhang.
\newblock Halo: Learning to prune neural networks with shrinkage.
\newblock In \emph{Proceedings of the 2021 SIAM International Conference on
  Data Mining (SDM)}, pages 558--566. SIAM, 2021.

\bibitem[Xue et~al.(2020)Xue, Zhou, Du, Dai, Xu, Zhang, and Cui]{xue2020deep}
Yuan Xue, Denny Zhou, Nan Du, Andrew~M. Dai, Zheen Xu, Kun Zhang, and Claire
  Cui.
\newblock Deep state-space generative model for correlated time-to-event
  predictions.
\newblock \emph{Proceedings of the 26th ACM SIGKDD International Conference on
  Knowledge Discovery and Data Mining}, pages 1552--1562, 2020.

\end{thebibliography}

\newpage
\section*{Appendix}

\section*{Model Details}
\subsection*{Generate RNN Variables}
The variables $\{\pmb{h}_t\}$ are estimated in a deterministic manner via a recurrent neural network with gated recurrent units. Recurrent neural network unfolds along the temporal horizon, making it a natural choice for the time series modeling. We write the update equation as follows:
\begin{align*} 
\pmb{h}_t &= \delta(\text{GRU}_{\pmb{\theta}}(\pmb{h}_{t-1}, \pmb{u}_t, \pmb{y}_{t-1}))
\end{align*}
where GRU$(\cdot)$ is gated recurrent unit update and $\delta(\cdot)$ is the delta-Dirac function. 

\subsection*{Inference of Global Shrinkage Variables}
The posterior for the global portion of the shrinkage prior $\pmb{g} = (\tau, c)$ are estimated conditional on the entirety of the observed data. We adopt two log-normal distributions for the approximate posterior of $\tau$. An additional log-normal will be used as the posterior of $c$. 
\begin{align*} 
    \widetilde{\tau}^2 &= \widetilde{\alpha}_{\tau} \widetilde{\beta}_{\tau} 
    &\widetilde{c}^2 \sim LN(\mu_{\pmb{\phi}, c}(\pmb{y}_{1:T}), \sigma_{\pmb{\phi}, c}(\pmb{y}_{1:T})) \\
    \widetilde{\alpha}_{\tau} &\sim LN(\mu_{\pmb{\phi}, \alpha, \tau}(\pmb{y}_{1:T}), \sigma_{\pmb{\phi}, \alpha, \tau}(\pmb{y}_{1:T})) 
    & \widetilde{\beta}_{\tau} \sim LN(\mu_{\pmb{\phi}, \beta, \tau}(\pmb{y}_{1:T}), \sigma_{\pmb{\phi}, \beta, \tau}(\pmb{y}_{1:T}))
\end{align*}
where once again $(\mu_{\pmb{\phi}, c}(\pmb{y}_{1:T})$, $\sigma_{\pmb{\phi}, c}(\pmb{y}_{1:T})$, $\mu_{\pmb{\phi}, \alpha, \tau}(\pmb{y}_{1:T})$, $\sigma_{\pmb{\phi}, \alpha, \tau}(\pmb{y}_{1:T})$, $\mu_{\pmb{\phi}, \beta, \tau}(\pmb{y}_{1:T})$, $\sigma_{\pmb{\phi}, \beta, \tau}(\pmb{y}_{1:T}))$ are all feed-forward neural networks. These neural networks takes average of all observed variables as inputs to allow for varying length sequences. Conditional on $(\widetilde{\tau}, \widetilde{c}, \pmb{\widetilde{\lambda}}_t)$, $\pmb{\widetilde{\tau}}_t^*$ can be estimated as follows: 
\begin{align*}
\pmb{\widetilde{\tau}}_t^{* 2} = \frac{\widetilde{c}^2\widetilde{\tau}^2}{\widetilde{c}^2 + \widetilde{\tau}^2 \pmb{\widetilde{\lambda}}_t^2}.
\end{align*}

\subsection*{Variational Inference / Forecasting Procedures}
Estimating the expectation term in the inference Equation remains a difficult task. To reduce the variance of the estimator, we utilize Stochastic Gradient Variational Bayes \citep{kingma2013auto, rezende2014stochastic} and the reparameterization trick. At each time step, we sample $q_{\pmb{\phi}}(\tau | \pmb{y}_{1:T})$, $ q_{\pmb{\phi}}(c | \pmb{y}_{1:T})$ and $q_{\pmb{\phi}}(\pmb{\lambda}_t | \pmb{z}_{t-1}, \pmb{h}_t)$ from the log-normal posterior distributions. However, instead of directly sampling from $q_{\pmb{\phi}}(\pmb{z}_t | \pmb{z}_{t-1}, \pmb{y}_t, \pmb{h}_t, \tau, c, \pmb{\lambda}_{t})$, we sample from auxiliary random variables $\pmb{\widetilde{\epsilon}}_t \sim N(\pmb{0}, \pmb{I})$ where $\pmb{I}$ is the identity matrix. We then obtain samples of the posterior through a transformation the auxiliary variables as follows: $\pmb{\widetilde{z}}_t = \pmb{\widetilde{\epsilon}}_t \odot \pmb{\widetilde{\sigma}}_{\pmb{z}, t} + \pmb{\widetilde{\mu}}_{\pmb{z}, t}$ where $\odot$ is element-wise product, $\pmb{\widetilde{\mu}}_{\pmb{z}, t} = \pmb{\mu}_{\pmb{\phi}, \pmb{\widetilde{z}}}(\pmb{\widetilde{z}}_{t-1}, \pmb{y}_t, \pmb{h}_t) \pmb{\widetilde{\tau}}_t^* \pmb{\widetilde{\lambda}}_t$ and
$\pmb{\widetilde{\sigma}}_{\pmb{z}, t} = \pmb{\sigma}_{\pmb{\phi}, \pmb{\widetilde{z}}}(\pmb{\widetilde{z}}_{t-1}, \pmb{y}_t, \pmb{h}_t) \pmb{\widetilde{\tau}}_t^* \pmb{\widetilde{\lambda}}_t$. By utilizing auxiliary random variables, we separate the source of randomness from the posterior parameters in which the gradients are required. This, in turn, reduces the variance of the gradient estimators.

During testing time, Monte Carlo samples will be utilized to generate forecasts from the model. Suppose the goal is to forecast $p(\pmb{y}_{(T+1):(T+p)} | \pmb{y}_{1:T}, \pmb{u}_{1:(T+p)})$ with $p$ denoting the forecasting horizon. First, we sample global variables from their approximate posterior $q_{\pmb{\phi}}(c, \tau | \pmb{y}_{1:T})$ using the sequence of known observed values. The sampled values will be used in all time-steps in the forecasting horizon.

Next, for each time-step from $1$ to $T$, we iteratively estimate all time-dependent variables using both the generative and the inference portions of the model. The latent variables will be sampled from the posterior distribution $q_{\pmb{\phi}}(\pmb{z}_t | \pmb{z}_{t-1}, \pmb{y}_t, \pmb{h}_t, \tau, c, \pmb{\lambda}_{t}) $ until the last known time-step. For each time-step $t$ in the forecasting horizon from $T+1$ to $T+p$, we iteratively update the model to obtain updates as follows: 
\begin{align*}
    \pmb{\widetilde{y}}_{T+t}  &\sim N(\pmb{A}_{\pmb{\theta}} \pmb{z}_{T+t}, \sigma_{\pmb{\theta}, \pmb{y}}(\pmb{z}_t)) \\
    \pmb{z}_{T+t} &= \pmb{z}_{T+t}^* \pmb{\widetilde{\tau}}_{T+t}^* \pmb{\widetilde{\lambda}}_{T+t} \\
    \pmb{z}_{T+t}^* &\sim N(\pmb{\mu}_{\theta, \pmb{z}}(\pmb{h}_{T+t}, \pmb{z}_{T+t-1}), \pmb{\sigma}_{\theta, \pmb{z}}(\pmb{h}_{T+t}, \pmb{z}_{T+t-1})) \\
    \pmb{h}_{T+t} &= \delta(\text{GRU}_{\pmb{\theta}}(\pmb{h}_{T+t-1}, \pmb{u}_{T+t}, \tilde{\pmb{y}}_{T+t-1}))
\end{align*}
where $\tilde{\pmb{y}}_{T+t-1}$ is the sampled response from the previous time-step. $\pmb{\widetilde{\lambda}}_{T+t}$ is sampled using the generator model and $\pmb{\widetilde{\tau}}_{T+t}^*$ is calculated using the inference model.

One note to highlight is that we sample the variables associated with the shrinkage prior from the inference model and the latent variables from the generative model. This is due to the fact that in the inference model, estimation of $\pmb{\widetilde{z}_t}^*$ requires $\pmb{y}_t$ which makes it infeasible during testing time. In contrast, we specifically designed the estimation for the approximate posterior of the shrinkage variables to not rely on $\pmb{y}_t$. This choice allows the shrinkage variables to be sampled from their approximate posteriors which provide more information. 

\section*{Model Details for Real World Datasets}
\subsection{Metrics Details}
For each of the models, we generate $n = 50$ number of Monte Carlo samples for each time-step in the forecasting horizon.  Suppose in the testing set, we wish to forecast $N$ series each with forecasting length $p$. The metrics are defined as follows:
\begin{align*}
    \text{ND} &= \frac{\sum_{n=1}^N \sum_{t=1}^p |y_{t, i} - \widetilde{y}_{t, i}|}{\sum_{n=1}^N \sum_{t=1}^p |y_{t, i}|} \\
    \text{RMSE} &= \frac{\sqrt{\frac{1}{Np}\sum_{n=1}^N \sum_{t=1}^p (y_{t, i} - \widetilde{y}_{t, i})^2}}{\frac{1}{Np}\sum_{n=1}^N \sum_{t=1}^p |y_{t, i}|}
\end{align*}
where $y_{t, i}$ is the true observed value for $i$th series in the testing set at forecast time-step $t$ and $\widetilde{y}_{t, i}$ is the median of the predicted samples from the model. These metrics present good evaluation metrics as they measure average deviation between the true values and the predicted values. 

\subsection*{Dataset Details}
\textit{Electricity} dataset contains hourly usage across many month. We utilize data from January 1, 2014 to September 1, 2014 for training/validation and data from September 1, 2014 to September 8, 2014 for testing. We evaluate the forecasting performances of the models over a forecast horizon of 48 hours with a learning period of 144 hours. Following \citep{salinas2020deepar}, each input series is standardized using a series dependent scale factor. This standardization process allows for efficient handling of drastically change scales inherent in the dataset.

The \textit{traffic} dataset contains 15 month of occupancy rate for San Francisco freeways. We utilize data before June 15, 2008 for training/validation and data from June 15, 2008 to June 22, 2008 for testing. Similar to electricity, we utilize a forecasting horizon of 48 hours with a learning period of 144 hours. The models are implemented via Pytorch using NVIDIA Tesla K80 GPU on Google Colab; it takes around 15 hours to train for \textit{electricity} and 12 hours to train for \textit{traffic}.

\subsection*{Hyperparameter Tuning Details}
This section will detail the hyperparameter selection process for DSSM-SH for both \textit{electricity} and \textit{traffic} datasets. For both datasets, we generate around 450,000 samples for training and 50,000 samples for validation. This is similar to amount of samples utilized in previous work \citep{lim2021temporal}. For \textit{electricity} dataset in particular, the observed data can be drastically different in scale for different series. To control scale handling, we use a weighted sampler \citep{salinas2020deepar} for the training set. The weighted sampler assigns a weight to each training sample based on the average magnitude of the response. We find in implementation that using this weighted sampler improves the training performances of the models. 

In terms of hyperparameter tuning for DSSM-SH, we tune 4 parameters: the dimension for the recurrent neural network (dim($\pmb{h}_t$)), the dimension for the latent variables (dim($\pmb{z}_t$)), number of layers for the recurrent neural network and the learning rate for the ADAM optimizer. The choices for each of the hyperparameters are given as follows:
\begin{itemize}
    \item dim($\pmb{h}_t$): 60, 80, 100, 120
    \item dim($\pmb{z}_t$): 10, 20, 30, 40
    \item Number of layers: 1, 2, 3
    \item Learning rate: $1\mathrm{e}{-3}$, $1\mathrm{e}{-4}$
\end{itemize}
We utilize random search in order to select the optimal hyperparameters using training/validation sets. The optimal parameters found for DSSM-SH for the \textit{electricity} dataset is dim($\pmb{h}_t$) = 80, dim($\pmb{z}_t$) = 30, number of layers = 2, learning rate = $1\mathrm{e}{-3}$. The resulting model has a total of 296808 number of parameters. The optimal parameters found for DSSM-SH for the \textit{traffic} dataset is dim($\pmb{h}_t$) = 80, dim($\pmb{z}_t$) = 20, number of layers = 2, learning rate = $1\mathrm{e}{-3}$. The resulting model has a total of 288668 number of parameters. The choices for the other hyperparameters are as follows: $\tau_0 = 1$, $c_a = 2$, $c_b = 1$. As seen, the model contains a high number of parameters, supporting the idea that adding shrinkage can be potentially very useful.  

\section*{Variational Inference Details}
In this section, we will give more details concerning the estimation of the cost function for our model. Given the details about the generative and the inference portion of the models seen in Section 3, we start with the ELBO as follows:
\begin{align*}
    L(\pmb{\theta}, \pmb{\phi}) &\equiv E_{q_{\pmb{\phi}}(\pmb{z}_{1:T}, \pmb{h}_{1:T}, \tau, c, \pmb{\lambda}_{1:T} | \pmb{y}_{1:T}, \pmb{u}_{1:T})} [\log p_{\pmb{\theta}}(\pmb{y}_{1:T} | \pmb{z}_{1:T}, \pmb{h}_{1:T}, \tau, c, \pmb{\lambda}_{1:T}, \pmb{u}_{1:T})] - \\
    & \; \; \; \; \; \; \; \; \; \text{KL}(q_{\pmb{\phi}}(\pmb{z}_{1:T}, \pmb{h}_{1:T}, \tau, c, \pmb{\lambda}_{1:T} | \pmb{y}_{1:T}, \pmb{u}_{1:T}) || p_{\pmb{\theta}}(\pmb{z}_{1:T}, \pmb{h}_{1:T}, \tau, c, \pmb{\lambda}_{1:T} | \pmb{y}_{1:T}, \pmb{u}_{1:T})) \\
    &\equiv L_E  - L_{KL}
\end{align*}
This loss has two terms which we will discuss individually. For clarity and ease of notation, we will denote the first term as $L_E$ and the second term as $L_{KL}$. We will start with a discussion for the KL term followed by discussion of the expectation term. We first perform the following factorization for $L_{KL}$:
\begin{align*}
   L_{KL} = \sum_{t=1}^T E_{q_{\pmb{\phi}}(\tau, c, \pmb{\lambda}_{t} | \pmb{y}_{t}, \pmb{h}_{t}, \pmb{u}_{t})}[&KL(q_{\pmb{\phi}}(\pmb{z}_{t} | \pmb{z}_{t-1}, \tau, c, \pmb{\lambda}_{t}, \pmb{y}_{t}, \pmb{h}_{t}, \pmb{u}_{1:T}) || \\
   & p_{\pmb{\theta}}(\pmb{z}_{t} | \pmb{z}_{t-1}, \tau, c, \pmb{\lambda}_{t}, \pmb{y}_{t},  \pmb{h}_{t}, \pmb{u}_{t}))] \\
   & + KL(q_{\pmb{\phi}}(\tau, c, \pmb{\lambda}_{t} | \pmb{z}_{t-1}, \pmb{y}_{1:T}, \pmb{u}_{t}, \pmb{h}_{t}) || p_{\pmb{\theta}}( \tau, c, \pmb{\lambda}_t))
\end{align*}
This factorization allow KL-divergence terms to be written as sum of individual KL-terms across time. The reason this factorization works is due to the structure we chose for the generative and inference models. Note that $\pmb{h}_{1:T}$ is deterministic and uses the same RNN for both the generative and inference portions. This choice leads to a KL-divergence of 0 for $\pmb{h}_{1:T}$ which simplify the derivation.

We will now break down each term in $L_{KL}$. First, in the generative model, the distributions $(\tau, c, \pmb{\lambda}_t)$ do not conditional on any other variables and are independent of one another. This allow us to drop the conditional for this term. In the inference model, conditional on $(\pmb{z}_{t-1}, \pmb{y}_{1:T}, \pmb{u}_{t}, \pmb{h}_{t})$, the approximate posterior of $(\tau, c, \pmb{\lambda}_{t})$ are independent of one another. This allow us to fully factorize $KL(q_{\pmb{\phi}}(\tau, c, \pmb{\lambda}_{t} | \pmb{z}_{t-1}, \pmb{y}_{1:T}, \pmb{u}_{t}, \pmb{h}_{t}) || p_{\pmb{\theta}}( \tau, c, \pmb{\lambda}_t))$ into separate terms and evaluated analytically. From the posterior and prior choices given in Section 3, this KL term has a closed form solution. This closed form solution relies the closed form KL-divergence formula between log-normal random variables and Gamma/inverse-Gamma random variables derived in \citet{louizos2017bayesian}.

Second, conditional on $(\pmb{z}_{t-1}, \tau, c, \pmb{\lambda}_{t}, \pmb{y}_{t}, \pmb{h}_{t}, \pmb{u}_{1:T})$, the distribution of $\pmb{z}_t$ is normal for both the generative and inference portion of the model. The conditional can be written as follows:
\begin{align*}
    q_{\pmb{\phi}}(\pmb{z}_{t} | \pmb{z}_{t-1}, \tau, c, \pmb{\lambda}_{t}, \pmb{y}_{t}, \pmb{h}_{t}, \pmb{u}_{t}) &\sim N(\pmb{\mu}_{\pmb{\phi}, \pmb{z}}(\pmb{z}_{t-1}, \pmb{y}_t, \pmb{h}_t) \pmb{\tau}_t^* \pmb{\lambda}_t, \pmb{\sigma}_{\pmb{\phi}, \pmb{z}}(\pmb{z}_{t-1}, \pmb{y}_t, \pmb{h}_t) \pmb{\tau}_t^* \pmb{\lambda}_t) \\
    p_{\pmb{\theta}}(\pmb{z}_{t} | \pmb{z}_{t-1}, \tau, c, \pmb{\lambda}_{t}, \pmb{y}_{t},  \pmb{h}_{t}, \pmb{u}_{t}) &\sim N(\pmb{\mu}_{\pmb{\theta}, \pmb{z}}(\pmb{z}_{t-1}, \pmb{y}_t, \pmb{h}_t) \pmb{\tau}_t^* \pmb{\lambda}_t, \pmb{\sigma}_{\pmb{\theta}, \pmb{z}}(\pmb{z}_{t-1}, \pmb{y}_t, \pmb{h}_t) \pmb{\tau}_t^* \pmb{\lambda}_t) 
\end{align*}
Suppose $\pmb{z}_t \in R^Q$ where $\pmb{z}_t = (z_{t, 1}, ..., z_{t, Q})$, the KL-divergence between the two terms can be calculated as follows:
\begin{align*}
&KL(q_{\pmb{\phi}}(\pmb{z}_{t} | \pmb{z}_{t-1}, \tau, c, \pmb{\lambda}_{t}, \pmb{y}_{t}, \pmb{h}_{t}, \pmb{u}_{t}) || p_{\pmb{\theta}}(\pmb{z}_{t} | \pmb{z}_{t-1}, \tau, c, \pmb{\lambda}_{t}, \pmb{y}_{t},  \pmb{h}_{t}, \pmb{u}_{t}))  \\
&= \sum_{i=1}^Q \log \frac{\sigma_{\pmb{\theta}, \pmb{z}, i}(\pmb{z}_{t-1}, \pmb{y}_t, \pmb{h}_t) \tau_{t, i}^* \lambda_{t, i}}{\sigma_{\pmb{\phi}, \pmb{z}, i}(\pmb{z}_{t-1}, \pmb{y}_t, \pmb{h}_t)\tau_{t, i}^* \lambda_{t, i}} - \frac{1}{2} \\
& \; \; \; \; \; \; \; \; \; \; \; \; \; \; \; \; + \frac{\sigma_{\pmb{\phi}, \pmb{z}, i}^2(\pmb{z}_{t-1}, \pmb{y}_t, \pmb{h}_t)\tau_{t, i}^{2 *} \lambda_{t, i}^{2} + (\mu_{\pmb{\phi}, \pmb{z}, i}(\pmb{z}_{t-1}, \pmb{y}_t, \pmb{h}_t) \tau_t^* \lambda_t - \mu_{\pmb{\theta}, \pmb{z}, i}(\pmb{z}_{t-1}, \pmb{y}_t, \pmb{h}_t) \tau_t^* \lambda_t)^2}{2\sigma_{\pmb{\theta}, \pmb{z}, i}^2(\pmb{z}_{t-1}, \pmb{y}_t, \pmb{h}_t)\tau_{t, i}^{2 *} \lambda_{t, i}^{2}} \\
&= \sum_{i=1}^Q \log \frac{\sigma_{\pmb{\theta}, \pmb{z}, i}(\pmb{z}_{t-1}, \pmb{y}_t, \pmb{h}_t)}{\sigma_{\pmb{\phi}, \pmb{z}, i}(\pmb{z}_{t-1}, \pmb{y}_t, \pmb{h}_t)} \\
& \; \; \; \; \; \; \; \; \; \; \; \; \; \; \; \; + \frac{\sigma_{\pmb{\phi}, \pmb{z}, i}^2(\pmb{z}_{t-1}, \pmb{y}_t, \pmb{h}_t)+ (\mu_{\pmb{\phi}, \pmb{z}, i}(\pmb{z}_{t-1}, \pmb{y}_t, \pmb{h}_t)  - \mu_{\pmb{\theta}, \pmb{z}, i}(\pmb{z}_{t-1}, \pmb{y}_t, \pmb{h}_t))^2}{2\sigma_{\pmb{\theta}, \pmb{z}, i}^2(\pmb{z}_{t-1}, \pmb{y}_t, \pmb{h}_t)} - \frac{1}{2}
\end{align*}
As seen, all terms involving the shrinkage variables $(\tau, c, \pmb{\lambda}_{1:T})$ cancels out in the KL-divergence. As a result,
\begin{align*}
    E_{q_{\pmb{\phi}}(\tau, c, \pmb{\lambda}_{t} | \pmb{y}_{t}, \pmb{h}_{t}, \pmb{u}_{t})}&[KL(q_{\pmb{\phi}}(\pmb{z}_{t} | \pmb{z}_{t-1}, \tau, c, \pmb{\lambda}_{t}, \pmb{y}_{t}, \pmb{h}_{t}, \pmb{u}_{1:T}) || p_{\pmb{\theta}}(\pmb{z}_{t} | \pmb{z}_{t-1}, \tau, c, \pmb{\lambda}_{t}, \pmb{y}_{t},  \pmb{h}_{t}, \pmb{u}_{t}))]  \\
    &= KL(q_{\pmb{\phi}}(\pmb{z}_{t} | \pmb{z}_{t-1}, \tau, c, \pmb{\lambda}_{t}, \pmb{y}_{t}, \pmb{h}_{t}, \pmb{u}_{1:T}) || p_{\pmb{\theta}}(\pmb{z}_{t} | \pmb{z}_{t-1}, \tau, c, \pmb{\lambda}_{t}, \pmb{y}_{t},  \pmb{h}_{t}, \pmb{u}_{t}))
\end{align*}
This once again allows for a closed-form solution which allows for effective evaluation of the KL-divergence. This solution is one of the key reasons behind many of the design choices for the generative/inference portions of the model. By choosing approximate conditional dependencies among variables, we allow for a closed-form solution for the KL-divergence in the loss. A closed-form solution leads to more accurate gradients and faster evaluations. 

Given a closed form solution for $L_{KL}$, we will now focus our attention to the first part of the loss: $L_E$. Similar to the first step in factorizing $L_{KL}$, we adopt the following factorization for $L_E$:
\begin{align*}
    L_E &= E_{q_{\pmb{\phi}}(\pmb{z}_{1:T}, \pmb{h}_{1:T}, \tau, c, \pmb{\lambda}_{1:T}| \pmb{y}_{1:T}, \pmb{u}_{1:T})} [\log p_{\pmb{\theta}}(\pmb{y}_{1:T} | \pmb{z}_{1:T}, \pmb{h}_{1:T}, \tau, c, \pmb{\lambda}_{1:T}, \pmb{u}_{1:T})] \\
    &= \sum_{t=1}^T E_{q_{\pmb{\phi}}(\pmb{z}_{t} | \pmb{z}_{t-1}, \pmb{h}_{t}, \tau, c, \pmb{\lambda}_{t}, \pmb{y}_{t}, \pmb{u}_{t})} [\log p_{\pmb{\theta}}(\pmb{y}_{t} | \pmb{z}_{t})]
\end{align*}
This factorization results directly from the choices we made in the generative model. Conditional on $\pmb{z}_t$, the generative distribution for $\pmb{y}_t$ is independent of over other variables in the model. To evaluate this expectation, we obtain samples from the approximate posterior $q_{\pmb{\phi}}(\pmb{z}_{t} | \pmb{z}_{t-1}, \pmb{h}_{t}, \tau, c, \pmb{\lambda}_{t}, \pmb{y}_{t}, \pmb{u}_{t})$ and evaluate the log predictive likelihood. 

Through our choices for the generative and inference portions of the model, we are able to obtain a time-based factorization of the loss. This allows the loss function to be calculated one-step at a time during training. With the closed-form KL-divergence terms, we do not need to sample any variables from the generative model in the training phase. We only need to sample from $q_{\pmb{\phi}}(\pmb{z}_{t} | \pmb{z}_{t-1}, \pmb{h}_{t}, \tau, c, \pmb{\lambda}_{t}, \pmb{y}_{t}, \pmb{u}_{t})$ at each time-step $t$ of the training process. For sampling, we utilize reparameterization trick as detailed above.

\end{document}